\documentclass[journal]{IEEEtran}
\ifCLASSINFOpdf
\else
\fi
\usepackage{multirow}
\usepackage{amsmath}
\usepackage{amssymb}
\usepackage{mathrsfs}
\usepackage{subfigure}
\usepackage{graphicx}
\usepackage{color}
\usepackage{times}
\usepackage{comment}
\usepackage{balance}
\usepackage{booktabs} 
\usepackage{bbm} 
\usepackage{bbding} 
\usepackage{algpseudocode} 
\usepackage{bm} 
\usepackage{threeparttable} 

\usepackage[linesnumbered,ruled]{algorithm2e} 

\newcommand*\hh{\mathcal{H}}
\newcommand*\sss{\mathcal{S}}
\newcommand*\ttt{\mathcal{T}}
\newcommand*\rr{\mathcal{R}}

\newcommand{\bb}[1]{\bm{#1}}
\newcommand{\ms}[1]{\mathcal{#1}}

\newcommand*\ie{{i.e.}}

\begin{document}
%
\title{Semi-supervised Domain Adaptive Structure Learning }

\author{\IEEEauthorblockN{Can Qin,
Lichen Wang,
Qianqian Ma, 
Yu Yin, Huan Wang, and
Yun Fu}





\thanks{Can Qin, Lichen Wang, Yu Yin, Huan Wang are with the Department of Electrical and Computer Engineering, Northeastern University, Boston, USA (Email: qin.ca@northeastern.edu, wanglichenxj@gmail.com, \{yin.yu1, wang.huan\}@northeastern.edu).}
\thanks{Qianqian Ma is with the Department of Electrical and Computer Engineering, Boston University, Boston, USA (Email: maqq@bu.edu).}
\thanks{Yun Fu is with the Department of Electrical and Computer Engineering, and Khoury College of Computer Science, Northeastern University, Boston, USA (Email: yunfu@ece.neu.edu).}
}

\maketitle


\begin{abstract}
Semi-supervised domain adaptation (SSDA) is quite a challenging problem requiring methods to overcome both 1) overfitting towards poorly annotated data and 2) distribution shift across domains. Unfortunately, a simple combination of domain adaptation (DA) and semi-supervised learning (SSL)  methods often fail to address such two objects because of training data bias towards labeled samples. In this paper, we introduce an adaptive structure learning method to regularize the cooperation of SSL and DA. Inspired by the multi-views learning, our proposed framework is composed of a shared feature encoder network and two classifier networks, trained for contradictory purposes. Among them, one of the classifiers is applied to group target features to improve intra-class density, enlarging the gap of categorical clusters for robust representation learning. Meanwhile, the other classifier, serviced as a regularizer, attempts to scatter the source features to enhance the smoothness of the decision boundary. The iterations of target clustering and source expansion make the target features being well-enclosed inside the dilated boundary of the corresponding source points. For the joint address of cross-domain features alignment and partially labeled data learning, we apply the maximum mean discrepancy (MMD) distance minimization and self-training (ST) to project the contradictory structures into a shared view to make the reliable final decision. The experimental results over the standard SSDA benchmarks, including DomainNet and Office-home, demonstrate both the accuracy and robustness of our method over the state-of-the-art approaches.

\end{abstract}


%
\IEEEpeerreviewmaketitle

\section{Introduction}
\IEEEPARstart{R}{ecently}, Deep Neural Networks (DNNs) have been extensively utilized in various tasks, e.g., image classification, text translation, and news recommendation~\cite{he2016deep,simonyan2014very,zhu2020incorporating, zheng2018drn}, ranging from computer vision to natural language understanding. However, despite the remarkable success, the DNNs heavily rely on huge amounts of annotated data for training, which are costly to obtain in practice.

\begin{figure}[t]
\centering
\scalebox{1}{\includegraphics[width=0.5\textwidth]{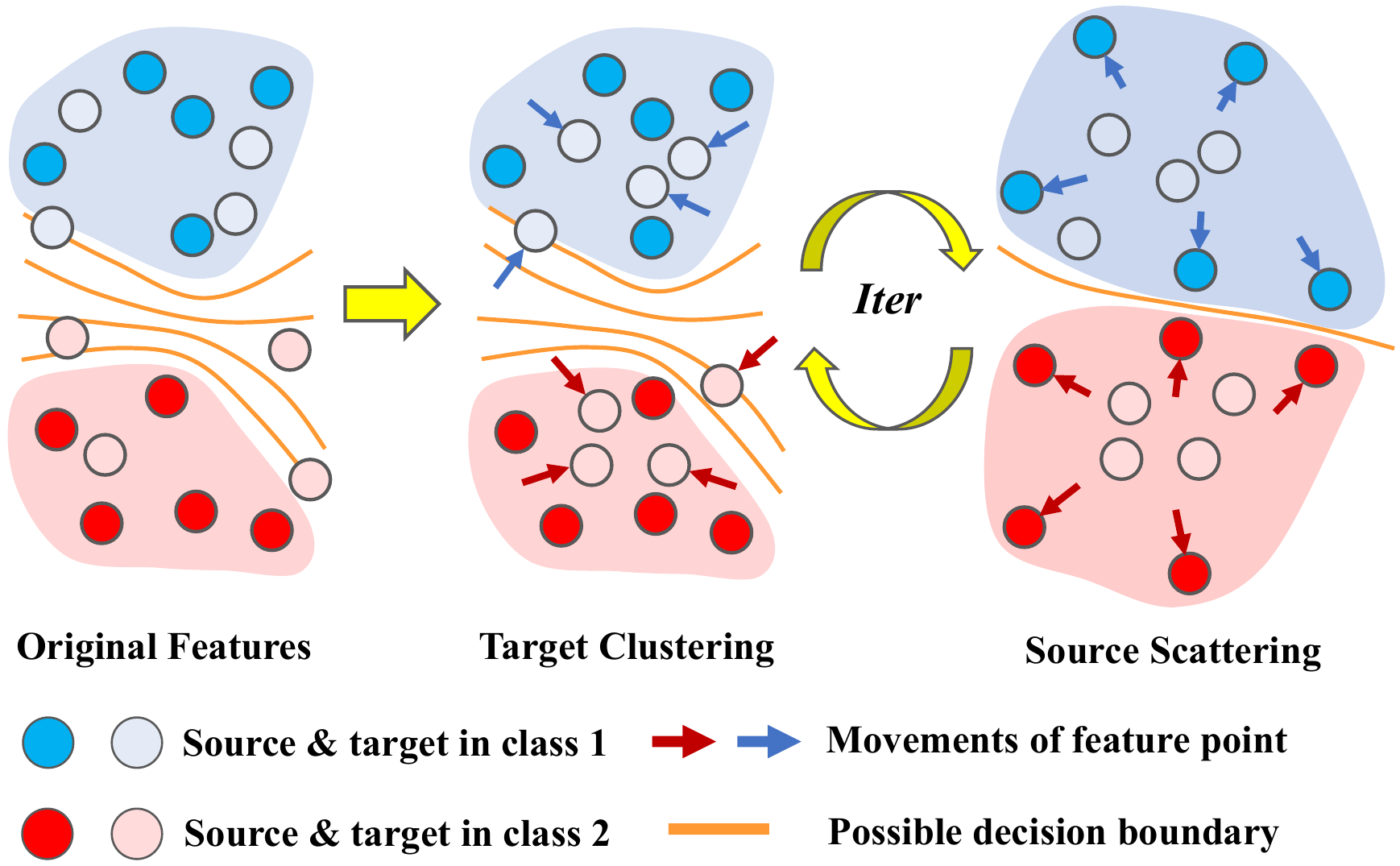}}
\caption{The illustration of our motivation. In Adaptive Structure Learning (ASL), there are alternative optimizations between source scattering and target clustering to explicitly align the cross-domain features. In the end, the target features are expected to be well enclosed within the expanded boundary of corresponding source features.
} \label{figs:concept}
\end{figure}

Domain Adaptation (DA)~\cite{peng2017visda,saito2019semi,sdm20_domain1,sdm20_domain2}, defined as transferring/adapting the model trained from the rich annotated dataset (i.e., source domain) to the label-scarce target domain, is one of the promising solutions to address such a data-deficiency challenge. In general, domain adaptation can be classified as unsupervised domain adaptation (UDA) and semi-supervised domain adaptation (SSDA) according to the access to target labels during training. This paper focuses on SSDA, with significantly better performance than UDA, given a tiny amount of labeled target samples. 

However, considerable distribution mismatches (i.e., domain gaps) between cross-domain data/features degenerate most conventional machine learning models built upon the distribution-sharing assumption. Feature alignment approaches attempt to match the cross-domain features explicitly by minimizing certain divergence or distances over the features projected in a shared feature space. How to define such a distance/divergence is the key to success. Several methods, such as Correlation Alignment (CORAL)~\cite{sun2015subspace}, Geodesic Flow Kernel (GFK)~\cite{gong2012geodesic,gopalan2011domain}, Maximum Mean Discrepancy (MMD)~\cite{long2013transfer}, have been developed. In recent years, implicit feature alignment methods have gained increasing popularity. Inspired by the development of Generate Adversarial Network (GAN)~\cite{goodfellow2014generative}, the typical adversarial training solution employs a zero-sum game between a domain classifier (\ie, discriminator) and a feature extractor to enforce the confusion of cross-domain features until the discriminator cannot tell any difference. However, the conditional distributions alignment still remains challenging due to the lack of target labels.


MME~\cite{saito2019semi} and DIRT-T~\cite{shu2018dirt} employed entropy loss minimization to group target domain features for conditional alignment by the assumption that well-clustered features are more discriminative. However, despite such a structure-wise regularization, the model is still strongly biased towards the source domain due to the inequality of labeled data across domains, inevitably leading to two negative impacts: 1) the decision boundary easily crosses the region of high-density target features mistakenly, and 2) the target domain features close to the decision boundary are likely driven to the wrong areas during alignment. Inspired by the noisy/soft labels commonly applied in weakly/semi-supervised learning, it is logical to assume that the decision boundary can be de-biased by the regularizer with slightly disturbed and scattered patterns. With this regard, UODA~\cite{qin2021contradictory} summarized that the ideal representation for SSDA should include conflicting aspects: 1) well-clustered target features and 2) scattered source features. Accordingly, this paper further extends this work as a more powerful adaptive structure learning framework for the semi-supervised domain adaptation problem, with the help of a generator and two classifier networks for different purposes.


However, UODA~\cite{qin2021contradictory} has two main limitations. Firstly, it overlooks the explicit alignment of cross-domain features, which provides a strong constraint to avoid mis-alignments. Secondly, it has not fully exploited the consistency between different data transformations, which is commonly adopted in SSL and has the great potential to learn more representative features. This paper attempts to extend UODA~\cite{qin2021contradictory} to address the both challenges. The minimization of MMD loss over the deep features helps to solve the first weakness. Such explicit alignment in the reproducing kernel Hilbert space (RKHS) has further supported the structure learning for precise conditional alignment. To solve the problem two, we utilize self-training to automatically achieve pseudo labels by enforcing the consistency between weakly and strongly transformed data. Such data transformations naturally bring the self-supervision for instance-wise regularization. The final scores on unlabeled data are jointly inferred by the two views/classifiers that provide complementary information for robust decisions. The whole pipeline can be trained in the end-to-end manner to ensure excellent cooperation between different modules, expected to overcome both domain gaps and data bias.


In summary, this paper has three major contributions as follows: 

\begin{itemize}
\item The implicit feature alignment based on contradictory structure learning might bring negative transfer due to conditional mismatch. This paper introduces the explicit alignment loss to assist the implicit one for the more precise alignment.

\item Instead of focusing on opposite structure learning, this paper pays more attention to reunite such contradictions, which are also crucial. To fulfill it, we take the self-training to enforce the consistency among the predictions of differently augmented data along with the opposite structure learning.

\item Extensive experiments on the popular benchmarks, such as Office-home~\cite{venkateswara2017deep} and DomainNet~\cite{peng2019moment}, demonstrate the advantage of our proposed approach over the direct baseline~\cite{qin2021contradictory} and other latest methods.\footnote{Our code can be seen at: https://github.com/canqin001/ASDA}
\end{itemize}


\section{Related Works}
\subsection{Unsupervised Domain Adaptation}
In conventional machine learning scenarios, the training and testing sets are supposed to share the same feature distributions. It is the essential assumption for most of the algorithms. However, as machine learning technologies extend to real-world applications, such as pandemic prediction~\cite{ma2020optimal}, and explore diverse data formats, the assumption is not always the truth anymore. Unsupervised Domain Adaptation (UDA) aims to adopt a model from the source domain to the target domain without supervision. Specifically, any labels from the target samples are inaccessible. Therefore, it is a more challenging task compared with the semi-supervised scenario. There are multiple UDA methods have been proposed~\cite{cao2018unsupervised,ganin2014unsupervised,khan2016adapting,wang2018visual,qin2019pointdan,qin2019generatively,dong2020cscl} in recent years which achieved promising results. These methods could be grouped into three categories, which are 1) source-target divergence metric-based methods, 2) generalization extension-based methods, and 3) constant term-based methods~\cite{ben2010theory}. The main strategy of the first category is projecting both source and target samples into a latent common space, where these samples could be well aligned by minimizing the divergence.


\begin{figure*}[t]
\centering
\scalebox{1}{\includegraphics[width=1\textwidth]{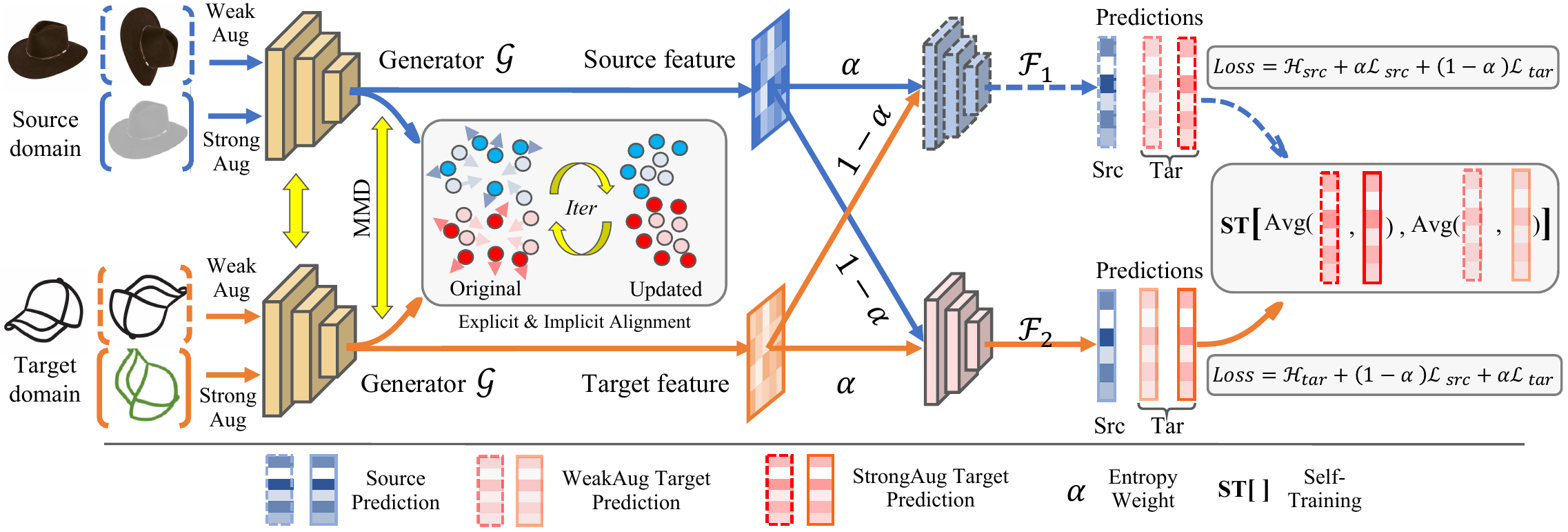}}
\caption{Illustration of the proposed Adaptive Structure Learning for Semi-supervised Domain Adaptation (ASDA) including three modules: 1) a deep feature encoder network  {$\mathcal{G}$}, 2) a source-scattering classifier network {$\mathcal{F}_1$}, and 3) a target-clustering classifier network {$\mathcal{F}_2$}. This paper takes a Convolutional Neural Network as {$\mathcal{G}$} and a two-layer Feed-forward Neural Network as either {$\mathcal{F}_1$} or {$\mathcal{F}_2$}. The raw data will be transformed into different formats as inputs according to the WeakAug and StrongAug operations. In this figure, both generators (in yellow) share the parameters for feature extraction. The two classifiers will take the features from the generator for classification. More details about the loss functions can be referred in Sec.~\ref{sec:method}.   
}\label{figs:framework}
\end{figure*}

Metric learning~\cite{jhuo2012robust, ma2020adversarial} used to be a popular solution for UDA.
As Deep Neural Networks have achieved considerable progress in recent decent, they are naturally explored in UDA to obtain representations. The straightforward strategy is projecting the source and target features into a common subspace where the divergence/shift across different domains would be minimized. \cite{sun2016deep} designed a correlation learning strategy that effectively aligns different domains. Maximum Mean Discrepancy is set as the metric for UDA~\cite{long2013transfer}. Although these methods achieved promising results, however, the distance metrics are fixed, which cannot well handle the various representations from different domains. Adversarial learning provides another strategy to evaluate the domain shift where a discriminator, as well as encoders, are deployed to competently obtain the domain-invariant representations (i.e., Gradient Reversal~\cite{ganin2014unsupervised}, ADDA~\cite{tzeng2017adversarial}). Specifically, CyCADA~\cite{pmlrhoffman18a} is an extension of the ADDA with the cycle consistency/translation~\cite{CycleGAN2017} to generate synthetic images consisting of both target styles and source content.



Although adversarial learning is effective in reducing the domain shift, the conditional distribution mismatch is still an unsolved problem. The multi-classifier framework is proposed in Tri-training~\cite{saito2017asymmetric}, which derives pseudo labels from multiple views. DIRT-T~\cite{shu2018dirt} fully explores the target features in both the feature space and label space. The density of the intra-class knowledge and the divergence of the inter-class knowledge is explored for tuning the classification boundaries. MCD~\cite{saito2018maximum} proposes another solution that deploys two classifiers to learn two different boundaries, where the differences of the boundaries are used in adversarial learning mechanisms to obtain the domain-invariant representations. However, due to the inaccessibility of target labels, it is tough to describe the conditional distribution of target features precisely. Thus, UDA methods have limited potentials for real-world applications.


\subsection{Semi-supervised Domain Adaptation}

Semi-supervised Domain Adaptation (SSDA) assumes that a few samples in the target domain are labeled, and the corresponding labels are accessible in the training stage. Due to this, it is also named Few-shot Domain Adaptation (FSDA) scenario. SSDA is a more practical setting in real-world applications since labeling a small number of samples is cost-efficient, and it could lead to considerable performance improvement. Furthermore, by exploring the supervision/label information, the distribution knowledge of the target domain could be fully explored in a more fine-grained way, which potentially improves the domain-adaptation performance. However, there are still drawbacks. Specifically, it is difficult to automatically balance the learning contribution between the source and target domains, which could cause the overfitting issue in the target domain and reduce the generalization ability and overall test performance.


There are a few works proposed for solving the challenges. \cite{tzeng2015simultaneous} explores the label correlations and transfers the learned semantic knowledge across the source and target domains. Soft label scores, as well as the matching loss, are utilized in this method. CCSA~\cite{motiian2017unified} effectively explores the latent label correlation knowledge from the target samples for conditional distribution matching. The feature and label consistency from source and target domains are jointly optimized via separation loss in a maximization manner. FADA~\cite{motiian2017few} is an extension of the CCSA method. Specifically, an extra adversarial learning module is designed, which includes multiple discriminators as well as a generator. The goal is to further stabilize and tune the learned domain invariant representations in different classes. MME~\cite{saito2019semi} jointly considers the source and target samples' predictions and maximizes the entropy of the predictions across domains. Moreover, the clustering strategy obtains the prediction confidence and gets more fine-grain entropy loss for encoder training. However, due to the considerable unbalanced sample numbers between source and target domains, it is still difficult to control the learned decision boundaries in the target domain.  UODA~\cite{qin2021contradictory} is proposed to address the above challenge by the opposite structure learning. However, consider the relatively small number of the labeled target samples, the unpredictable sample distributions (e.g., density and divergence) could further confuse the classifiers in the target domain. CDAC~\cite{li2021cross} is the latest technique that employs both adaptive adversarial clustering and pseudo labeling with a set of data augmentations. This paper further improves CDAC based on the two classifiers/views, which provide complementary information for making the robust decision. Furthermore, the confusion of the classifiers in UODA can also be corrected by the explicit alignment cooperated with the implicit structure learning.


\section{Proposed Approach}
\label{sec:method}

\begin{table}[t]
\centering
\caption{NOTATIONS of Symbols}\label{tab:symbols}

\scalebox{0.9}{
\begin{tabular}{ c  l }
\toprule
\textbf{Symbol} &  \textbf{Descriptions} ~~~~~~~~~~~~~~~~~~~~~~~~~~~~~~~~  \\
\midrule
$x^s_i$, $y^s_i$ & the $i$-th source image data and groundtruth label\\
$x^t_i$, $y^t_i$ & the $i$-th labeled target image data and groundtruth label\\
$x^u_i$ & the $i$-th unlabeled target image data\\
$f_{i}$ & the deep feature of the $i$-th input data\\
$\mathcal{S}$, $\mathcal{T}$, $\mathcal{U}$ & the sets of source, labeled and unlabeled target domain\\
$\mathcal{G}(\cdot)$, $\mathcal{F}_1(\cdot), \mathcal{F}_2(\cdot)$ & encoder, source-scattering and target-clustering classifier \\
$d$, $K$ & dimensions of feature space and label space\\
$N_s$, $N_t$, $N_u$ & source, labeled and unlabeled target data point numbers\\
$\alpha$, $\beta$, $\lambda$  & loss weights\\
$x'$, $x''$ & weakly augmented and strongly augmented data\\

$\kappa(\cdot)$ & kernel function for MMD computation \\
$\sigma(\cdot)$ & softmax function \\
$\mathbbm{1}_{[\cdot]}$ & identity function \\

${\Theta}_\mathcal{G}$ & parameters of feature encoder network $\mathcal{G}$ \\

${\Theta}_{\mathcal{F}_1}$ & parameters of source scattering classifier $\mathcal{F}_1$\\

${\Theta}_{\mathcal{F}_2}$ & parameters  of target clustering classifier $\mathcal{F}_2$ \\

$p_1(\cdot)$, $p_2(\cdot)$ & predictions of classifier $\mathcal{F}_1$ and $\mathcal{F}_2$ \\

$\Gamma(\cdot)$ & average of two classifiers' predictions \\
\bottomrule
\end{tabular}
}
\end{table}

\subsection{Problem Setting and Motivation.}

In SSDA, the model is expected to be well generalized on the target domain with fully labeled source samples and partially labeled target samples. To formulate this, suppose ${\mathcal{S} = \{ \bm{x}^s_i, y^s_i \}^{N_s}_{i=1}}$ is the set of the source domain data, ${\mathcal{T} = \{ \bm{x}^t_i, y^t_i \}^{N_t}_{i=1}}$ is the set of labeled target data, where $N_s=|S|$ and $N_t=|T|$ is the size of the dataset $\ms S$ and $\ms T$, respectively. $\bm{x}_i^s$ and $\bm{x}_i^t$ represents the labeled source and target images, respectively, and $y^s$, $y^t$ denote the corresponding labels.  Apart from the labeled data, the unlabeled target image set ${\mathcal{U} = \{\bm{x}^u_i\}^{N_u}_{i=1} }$, on which the model is expected to be well generalized, is composed of $N_u$ images where $N_u=|\ms U|$ and $N_u \gg N_t$. Due to the distributions mismatch across domains, i.e., $p({\bm{x}}^s)\not=p({\bm{x}}^u)$, and imbalance between labeled and unlabeled data, i.e., $N_u \gg N_t$, the model usually performs poorly if trained only by the supervised learning method.

As shown in Fig.~\ref{figs:framework}, there are three modules for our proposed model: 1) the siamese feature extractor networks $\mathcal{G}(\cdot)$, 2) the source-scattering classifier $\mathcal{F}_{1}(\cdot)$, and 3) the target-clustering classifier $\mathcal{F}_{2}(\cdot)$. Specifically, the cross-domain data would be projected into a shared latent space with the help of siamese feature extractor networks. The source-scattering is designed to disturb the source features as the structure regularization. The target-clustering classifier implicitly groups the target features expected to be enclosed by the corresponding stretched source features. Next, We will explain in details of the proposed framework as well as the training procedure.

\subsection{Supervised Training.}
Our model requires minimizing the empirical risk on both the labeled source and target sets as conventional DA methods. Therefore, we take a Convolutional Neural Network (CNN) as the feature extractor network $\mathcal{G}(\cdot)$ to map the image $\bm{x}$ into the $d$-dimensional feature $ \bm{f}\in\mathbb{R}^{d}$ as:
\begin{equation}\label{eq:feat_encoding}
    \bm{f}=\mathcal{G} ( \bm{x}, {\Theta}_\mathcal{G}), 
\end{equation}
where $\bm{x}$ can come from either domains, and ${{\Theta}_\mathcal{G}}$ represents the parameter set of $\mathcal{G}(\cdot)$. Then, the classifiers $\mathcal{F}_1(\cdot)$ and $\mathcal{F}_2(\cdot)$ would classify the input feature $\bm{f}$ into $K$ categories as following:
\begin{align} \label{eq:classification}
     \bm{p}_1(y|\bm{x})=&\sigma\left(\ms F_1( \mathcal{G}(\bm{x}), {\Theta}_{\mathcal{F}_1})\right), \nonumber \\
     \bm{p}_2(y|\bm{x})=&\sigma(\ms F_2( \mathcal{G}(\bm{x}), {\Theta}_{\mathcal{F}_2})),
\end{align}
where $\sigma(\cdot)$ is the softmax function and ${ \bm{p}_1({y}|\bm{x})}$ and ${ \bm{p}_2({y}|\bm{x})} \in \mathbb{R}^{K}$ are the $K$-dimensional softmax scores  of classifiers $\mathcal{F}_1(\cdot)$ and $\mathcal{F}_2(\cdot)$, which are parameterized by the ${\Theta}_{\mathcal{F}_1}$ and ${\Theta}_{\mathcal{F}_2}$. The two classifiers, which are Feed-forward Neural Networks (FFN), are set as the same architecture with different parameters.

For model training, the empirical loss is composed of two items: source-domain supervision $\mathcal{L}_{src}$ and the target-domain supervision $\mathcal{L}_{tar}$ formulated as:
\begin{align}
\mathcal{L}_{src} = - &\mathbb{E}_{(\bm{x}^s,y^s)  \sim \mathcal{S}} \left[ \sum_{k=1}^{K}  \mathbbm{1}_{k=y^s}{\mathrm{\log}{\left( {p}_1({y}=y^s| \bm{x}^s) \right)}} \right],\label{eq:src_loss} \\
\mathcal{L}_{tar}= - &\mathbb{E}_{(\bm{x}^t,y^t)  \sim \mathcal{T}} \left[ \sum_{k=1}^{K}  \mathbbm{1}_{k=y^t}{\mathrm{\log}{\left( {p}_2({y}=y^t| \bm{x}^t)\right)} }\right] . \label{eq:tar_loss}
\end{align}

The discrepancy between the two classifiers plays an important role for model training. Therefore, to enforce $\Theta_{\mathcal{F}_1} \neq \Theta_{\mathcal{F}_2}$, we apply different weights on both supervision losses $\mathcal{L}_{src}$ and $\mathcal{L}_{tar}$ where $\mathcal{F}_1(\cdot)$ takes $\alpha \mathcal{L}_{src} + (1- \alpha)\mathcal{L}_{tar} $, and $(1- \alpha) \mathcal{L}_{src} +  \alpha \mathcal{L}_{tar} $ is for $\mathcal{F}_2(\cdot)$ with $\alpha$ assigned as 0.75.

\subsection{Features Alignment.}
To bridge the domain gap, this paper proposes the alignment of the features in both explicit and implicit ways. For explicit alignment, we directly minimize the MMD loss over the cross-domain features, and learning the opposite feature structures can be regarded as the implicit alignment.

\textbf{Implicit Alignment:}
As separate views, the two classifiers are assigned with different purposes to learn the easy-to-adapt structure expected to have the dense target features enclosed by the sparse source features. The minimization of conditional entropy on output softmax scores enforces the high-confident predictions for features clustering. To fulfill such an object, this paper applies the entropy loss to measure features' closeness. Therefore, the conditional entropy over the unlabeled target domain can be denoted as:




\begin{equation}
   H_{tar}= - \mathbb{E}_{\bm{x}^u  \sim \mathcal{U}}
   \left[ \sum_{k=1}^{K}  {p_2}({y}=k| \bm{x}^u)
    {\mathrm{\log} \ {{p_2}({y}=k| \bm{x}^u)}} \right] , \label{eq:tar_ent}
\end{equation}
where ${p_2}({y}=k| \bm{x}^u)$ represents the possibility of data $\bm{x}^u$ as the class $k$. 

Feature scattering can be simply seen as the reverse of feature clustering. To this end, we will also take the conditional entropy loss $H_{src}$ to implement the source feature expansion:
\begin{equation}
   H_{src}= -\mathbb{E}_{\bm{x}^s  \sim \mathcal{S}} \left[\sum_{k=1}^{K}  {p_1}({y}=k| \bm{x}^s)
    {\mathrm{\log} \ {{ p_1}({y}=k| \bm{x}^s)}}\right], \label{eq:src_ent}
\end{equation}
where $H_{src}$ would be optimized in the reverse way as $H_{tar}$ for learning the opposite structure.


\textbf{Explicit Alignment:} 
It is hard to say that the cross-domain features are well aligned only with the implicit alignment adopted in UODA~\cite{qin2021contradictory} and MME~\cite{saito2019semi}. Instead, such a structure-wise constraint cannot precisely ensure the match between the corresponding feature groups across domains. Therefore, narrowing certain divergence/distance of cross-domain features plays a crucial role in closing the relaxing gap inherited from implicit alignment. In this paper, we minimize the MMD~\cite{borgwardt2006integrating,long2013transfer} loss to align cross-domain features explicitly as:

\begin{align}
\mathcal{L}_{align} = & \frac{1}{N_s  N_s}\sum_{i,j=1}^{N_s}\kappa (\bm{f}_i^s,\bm{f}_j^s) \nonumber+\frac{1}{N_s  N_u}\sum_{i,j=1}^{N_s,N_u}\kappa (\bm{f}_i^s,\bm{f}_j^t) \\
&+\frac{1}{N_u  N_u}\sum_{i,j=1}^{N_u}\kappa (\bm{f}_i^t,\bm{f}_j^t), \label{eq:mmd_loss}
\end{align}
where $\kappa(
\cdot)$ is a kernel function and we apply the combination of Linear and Radial Basis Function (RBF) kernel in our model. $\bm{f}^s$ and $\bm{f}^t$ are the deep features of source domain and target domain, respectively.

\subsection{Views-consistent Self-training.}
Although the domain alignment is proposed to address the distribution shift, the over-fitting problem is still unsolved yet. Inspired by the techniques from semi-supervised learning~
\cite{sohn2020fixmatch}, the variant augmented data are crucial to enriching the training set, which can be naturally applied to regularize the decision boundary towards smoothing. Therefore, this paper also employs the self-training strategy where both weakly augmented data $\ms U' = \{\bm{x}_i^{u'}\}_{i = 1}^{N_u'}$ and strongly augmented data $\ms U''=\{\bb{x}_i^{u''}\}^{N_{u}''}_{i=1}$ can be automatically obtained as $\ms U' = WeakAug(\ms U)$ and $\ms U'' = StrongAug(\ms U)$. The weakly augmenting includes random horizontal flip and random crop, and the strong augmentation randomly selects two transformations from the ten choices~\cite{sohn2020fixmatch}. The hard pseudo labels are obtained by the weakly augmented data with the highest confidence among all the classes, i.e., $\hat{y}'=\mathrm{argmax}(\Gamma(x^{u'}) )$ where $\Gamma(\cdot) = \frac{p_1(\cdot) + p_2(\cdot)}{2}$ denotes the average predictions of two classifiers. Such pseudo labels are applied as the ground truth for self-training:

\begin{align}
\mathcal{L}_{st} = & -  \mathbb{E}_{\bm{x}^{u'}  \sim \mathcal{U'},  \bm{x}^{u''}  \sim \mathcal{U''}} \left[ \sum_{k=1}^{K}\mathbbm{1}_{\Gamma(y=k|x^{u'})>\tau}{\hat{y}'} \right. \nonumber\\
 & \log \left(\Gamma(y=k|x^{u''})\right) \Bigg], \label{eq:loss_st}
\end{align}


where a threshold $\tau$ is applied to filter the high-confident pseudo labels and $ \mathbbm{1}_{[\cdot]}$ is the identity function. For convenient denotation, the weakly augmented unlabeled target set $\mathcal{U}'$ is equal to the original unlabeled target set $\mathcal{U}$ in the practice.

\subsection{Overall Training.}

In this paper, there are three groups of parameters, i.e, ${\Theta}_{\mathcal{F}_{1}}$, ${\Theta}_{\mathcal{F}_{2}}$ and ${\Theta}_\mathcal{G}$, to be optimized based on multiple loss functions. The back-propagation of overall gradients starts from the two classifiers, and the cross-entropy (Eq.~\ref{eq:src_loss} and Eq.~\ref{eq:tar_loss}) is an important target of all the objectiveness.  We firstly apply the different weights on the task losses to enforce the difference between ${\Theta}_{\mathcal{F}_{1}}$ and ${\Theta}_{\mathcal{F}_{2}}$. Then, we maximize the target-domain entropy loss (Eq.~\ref{eq:tar_ent}) to update the class-wise prototypes and minimize the source-domain entropy loss (Eq.~\ref{eq:src_ent}) to make source features slightly gathered. Moreover, the self-training loss (Eq.~\ref{eq:loss_st}) is also necessary for model training. Therefore, the overall training objectiveness for the two classifiers can be written as:

\begin{align}
    {\Theta}_{\mathcal{F}_{1}}^{*} =&\mathop {\arg \min} \limits_{{\Theta}_{\mathcal{F}_{1}}} (1- \alpha)\mathcal{L}_{tar} + \alpha \mathcal{L}_{src} + \beta H_{src} + \mathcal{L}_{st},\label{eq:f1_opti}\\
    {\Theta}_{\mathcal{F}_{2}}^{*} =&\mathop {\arg \min} \limits_{{\Theta}_{\mathcal{F}_{2}}} (1- \alpha) \mathcal{L}_{src} +  \alpha \mathcal{L}_{tar} - \lambda H_{tar} + \mathcal{L}_{st}, \label{eq:f2_opti}
\end{align}
where $\lambda$ and $\beta$ are loss weights to balance the influence of conditional entropy loss and supervision loss. The minimization of conditional entropy loss drives the features away from the decision boundary to implicitly cluster features. In turn, maximizing the conditional entropy loss is the reverse operation which leads to the features scattering. The self-training loss helps reunite the opposites to learn the consistent predictions on the augmented data, which serves as the views-consistent regularization for robust decisions.

\begin{algorithm}[t]
\label{alg:overall_training}
\caption{Overall Training}
\textbf{Input}{: Labeled source set $\mathcal{S}$, labeled target set $\mathcal{T}$ and unlabeled target set $\mathcal{U}$ (i.e., $\mathcal{U'}$). Strongly augmented unlabeled target set $\mathcal{U''}$.  The number of training epochs \({T}\).  The hyper-parameters $\alpha$, $\beta$, $\tau$ and $\lambda$.}. \\
\textbf{Initialize}: The sets of parameters of encoder network, source scattering classifier and target clustering classifier: ${\Theta}_\mathcal{G}^{0}$, $ {\Theta}_{\mathcal{F}_1}^{0}$ and $ {\Theta}_{\mathcal{F}_2}^{0}$. \\

\For{t = 1 $\sim$  T } 
{
 \textcolor{blue}{\textit{\# Loss Computation}} \\
    Get the cross-domain features from encoder network with Eq.~\ref{eq:feat_encoding}; \\
    Compute the MMD loss based on the kernel function of Eq.~\ref{eq:mmd_loss}; \\
    Compute the cross-entropy loss of all the labeled data on both classifiers with Eq.~\ref{eq:classification}. \\
    Compute the source entropy and target entropy based on  Eq.~\ref{eq:src_ent} and Eq.~\ref{eq:tar_ent}. \\
    Compute the self-training loss based on  Eq.~\ref{eq:loss_st}. \\
    \textcolor{blue}{\textit{\# Parameters Updating}} \\
    {
    Update $ {\Theta}_{\mathcal{F}_1}^{t-1}$ to  $ {\Theta}_{\mathcal{F}_1}^{t}$ by Eq.~\ref{eq:f1_opti}. \\ 
    Update $ {\Theta}_{\mathcal{F}_2}^{t-1}$ to  $ {\Theta}_{\mathcal{F}_2}^{t}$ by Eq.~\ref{eq:f2_opti}. \\
    Update ${\Theta}_\mathcal{G}^{t-1}$ to ${\Theta}_\mathcal{G}^{t}$ by Eq.~\ref{eq:g_opti}.
    }
    

}

\textbf{Output}: ${\Theta}_\mathcal{G}^{*}$ $\leftarrow$ ${\Theta}_\mathcal{G}^{T}$, 
$ {\Theta}_{\mathcal{F}_1}^{*}$ $\leftarrow$
$ {\Theta}_{\mathcal{F}_1}^{T}$ and 
$ {\Theta}_{\mathcal{F}_2}^{*}$ $\leftarrow$ 
$ {\Theta}_{\mathcal{F}_2}^{T}$.
\end{algorithm}

To progressively disperse source features and cluster target features, the feature encoder network $\mathcal{G}(\cdot)$ is trained by the reversal of conditional entropy loss, where we attempt to minimize the target entropy loss and maximize the source entropy loss. Apart from the task loss and self-training loss inherited from the classifiers, we also take the MMD loss for the explicit alignment  (Eq.~\ref{eq:mmd_loss}). The overall loss for training the feature encoder network is formalized as below:
\begin{align}\label{eq:g_opti}
{\Theta}_\mathcal{G}^{*} =&\mathop {\arg \min} \limits_{{\Theta}_\mathcal{G}} \mathcal{L}_{src} +  \mathcal{L}_{tar} - \beta H_{src} + \lambda H_{tar} + \mathcal{L}_{st} + \mathcal{L}_{align}.
\end{align}

The whole framework is trained in an end-to-end manner with the help of gradient reversal~\cite{ganin2016domain} for adversarial training and would continue to loop until reaching the ending epochs. The ablation studies on different components can be referred to Sec.~\ref{sec:analysis}.

\subsection{Theoretical Insights}
Next, we provide some theoretical insights behind our proposed framework, especially for the design of adaptive structure learning. According to the conclusion in \cite{ben2010theory}, the risk for the target domain can be bounded by the risk for the source domain and the domain divergence as follows,
\begin{align}
    \forall h\in H, \quad \rr_\ttt(h)\leq \rr_\sss(h)+\frac{1}{2}d_{\hh}(\sss,\ttt)+\delta,
\end{align}
where $\sss$, $\ttt$ represents the source domain and target domain, respectively, $\rr_\ttt(h)$, $\rr_{\sss}(h)$ represents the expected risk on domain $\ttt$ and $\sss$, respectively, $d_{\hh}(p,q)$ is the $\hh$-distance between distribution $p$ and $q$,  $\delta$ is a constant which is related to the error of a perfect hypothesis on both domains. In this case, if we can decrease the value of $d_{\hh}(\sss,\ttt)$ via training the domain classifiers and the feature generators, we can decrease the risk on the target domain. Now, we will show in details how we achieved this. As $d_{\hh}(\sss,\ttt)$ is defined as: 
\begin{align}\label{divergence}
    d_\hh(\sss,\ttt)=2\sup_{h\in \hh} \left| \Pr_{\bb f^s \sim p }[h(\bb f^s)=1]- \Pr_{\bb f^t \sim q }[h(\bb f^t)=1]\right|,
\end{align}
where $\bb f^s$ and $\bb f^t$ represents the features extracted from domain $\sss$ and domain $\ttt$ respectively. In our framework, we use the entropy function $H(\cdot)$ to train the parameters of $\ms F_1(\cdot)$, $\ms F_2(\cdot)$ and $\ms G(\cdot)$. Though the entropy function is not the usual classification loss, our framework can be seen as minimizing divergence (\ref{divergence}) via adversarial training strategy on target domain and source domain. Suppose $h$ is a binary classifier as follows:
\begin{align}
    h(f)=\begin{cases}
    1 & {\rm if}\quad  H(\ms F_i(\bb f)) \geq \gamma\\
    0 & {\rm otherwise}
    \end{cases},
\end{align}
where $i=1,2$, $\gamma$ is a threshold. To facilitate the analysis, we assume the output of the classifiers $\ms F_1(\cdot)$ and $\ms F_2(\cdot)$ are the conditional probabilities. Then \eqref{divergence} can be rewritten as:
\begin{align}\label{divergence2}
    d_\hh(\sss,\ttt) &\approx 2\sup_{\ms F_1, \ms F_2} \left| \Pr_{\bb f^s \sim p }[H(\ms F_1(\bb f^s))\geq \gamma] \right.\nonumber\\
    &\left.\quad - \Pr_{\bb f^u \sim q }[H(\ms F_2(\bb f^u))\geq \gamma]\right| \nonumber \\
    & = 2\sup_{\ms F_1, \ms F_2} \left( \Pr_{\bb f^u \sim q }[H(\ms F_2(\bb f^u))\geq \gamma] \right. \nonumber\\
    &\quad \left.-\Pr_{\bb f^s \sim p }[H(\ms F_1(\bb f^s))\geq \gamma]\right), 
\end{align}
where the approximate equality is because we are using unlabeled data here. As the number of the unlabeled samples on target domain is much larger than the labeled ones, we can use the probability on unlabeled samples to replace the probability on whole target domain. The equality above is due to the assumption that $\Pr_{\bb f^u \sim q }[H(\ms F_2(\bb f^u))\geq \gamma]\geq \Pr_{\bb f^s \sim p }[H(\ms F_1(\bb f^s))\geq \gamma]$. This can be easily achieved as the labels of all the samples in the source domain are available, which implies that we can make the corresponding entropy be $0$. Next,  we replace $\sup$ with $\max$ in (\ref{divergence2}), then we can get

\begin{align*}
     d_\hh(\sss,\ttt)&\approx  2\max_{\ms F_1, \ms F_2} \left( \Pr_{\bb f^u \sim q }[H(\ms F_2(\bb f^u))\geq \gamma] \right.\\
     &\quad- \left. \Pr_{\bb f^s \sim p }[H(\ms F_1(\bb f^s)) \geq \gamma]\right ) \nonumber \\ 
     & = 2\min_{\ms F_1, \ms F_2} \left(-\Pr_{\bb f^u \sim q }[H(\ms F_2(\bb f^u))\geq \gamma] \right.\\
     &\left. \quad+\Pr_{\bb f^s \sim p }[H(\ms F_1(\bb f^s))\geq \gamma]\right ) \nonumber \\
    & = -2 \min_{\ms F_2} \Pr_{\bb f^u \sim q }[H(\ms F_2(\bb f^u))\geq \gamma] \\
    &\quad+ 2\min _{\ms F_1} \Pr_{\bb f^s \sim p }[H(\ms F_1(\bb f^s))\geq \gamma],
\end{align*}
which matches with the update rules (\ref{eq:f1_opti}), (\ref{eq:f2_opti}) in our framework. Intuitively, the training of classifers $\ms F_1$ and $\ms F_2$ can be seen as approximating the divergence $d_\hh(\sss,\ttt)$, which defines an upper bound of the target domain risk. Moreover, we aim to minimize the divergence with respect to the features $\bb f^s$ and $\bb f^u$ to bound the risk on $\ttt$: \begin{align}\label{problem}
 &\min_{\bb f^s,\bb f^u}\left\{ \max_{\ms F_1, \ms F_2} \left( 2 \Pr_{\bb f^u \sim q }[H(\ms F_2(\bb f^u))\geq \gamma] \nonumber \right.\right.\\
 & \left.\left.-2\Pr_{\bb f^s \sim p }[H(\ms F_1(\bb f^s))\geq \gamma]\right )\right\},
\end{align}
where finding optimal $\bb f^s$ and $\bb f^u$ is equivalent to finding the optimal feature extractor $\ms G(\cdot)$, which corresponds to the update of  \eqref{eq:g_opti} in our model.
To summarize, our maximum training process with classifier $\ms F_1$ and $\ms F_2$ can be seen as measuring the domain divergence, while minimum training process with generator $\mathcal{G}$ can be seen as minimizing the divergence. In this case, we can effectively reduce the risk on the target domain $\ttt$. The discussions of MMD for DA can be referred to in many previous works~\cite{long2015learning,wang2020rethink}, and we will leave the theoretical analysis of ST loss in future works.

\section{Experiments}

\subsection{Experiments Setup.}

\textbf{Implementation.} For the evaluation of different architectures, we choose both the VGG16~\cite{simonyan2014very} and ResNet34~\cite{he2016deep} as the backbones of the feature encoder network $\mathcal{G}(\cdot)$. The two classifiers $\mathcal{F}_1(\cdot)$ and $\mathcal{F}_2(\cdot)$ utilize the same architecture Feed-forward Network (FFN) with two layers randomly initialed. For model optimization, we take the momentum Stochastic Gradient Descent (SGDM) as the optimizer on PyTorch~\cite{NEURIPS2019_9015}. The learning rate is assigned as $0.01$ with the momentum as $0.9$ and weight decay as $0.0005$. The loss weights $\alpha$, $\beta$ and $\lambda$ are assigned as $0.75$, $0.1$, and $0.1$ respectively. The strong data augmentation strategy consists of a sequence of two operations randomly picked from the ten following~\cite{sohn2020fixmatch}.



\begin{table*}[t]
\begin{center}
\caption{ {Quantitative results (\%) on the DomainNet~\cite{peng2019moment} under ResNet-34~\cite{he2016deep}. }}
\label{tab:quant_domainnet}
\scalebox{0.925}{
\begin{threeparttable}
 \centering
  \begin{tabular}{|c|cccccccccccccc|cc|}
   \hline 
\multicolumn{17}{|c|}{ResNet-34}  \\\hline
   \multirow{2}{*}{Methods} & \multicolumn{2}{c}{R$\rightarrow$C} & \multicolumn{2}{c}{R$\rightarrow$P} & \multicolumn{2}{c}{P$\rightarrow$C} & \multicolumn{2}{c}{C$\rightarrow$S} & \multicolumn{2}{c}{S$\rightarrow$P} & \multicolumn{2}{c}{R$\rightarrow$S} & \multicolumn{2}{c}{P$\rightarrow$R} & \multicolumn{2}{|c|}{Avg}\\
   
   &{1$_{shot}$} &{3$_{shot}$} &{1$_{shot}$} &{3$_{shot}$} &{1$_{shot}$} &{3$_{shot}$} &{1$_{shot}$} &{3$_{shot}$} &{1$_{shot}$} &{3$_{shot}$} &{1$_{shot}$} &{\small{3-shot}} &{1$_{shot}$} &{3$_{shot}$} &{1$_{shot}$} &\multicolumn{1}{c|}{3$_{shot}$} \\
   \hline
   
\multicolumn{1}{|c|}{S+T} &55.6&60.0  &60.6&62.2   &56.8&59.4  &50.8&55.0  &56.0&59.5  &46.3&50.1 &71.8&73.9 &56.9&\multicolumn{1}{c|}{60.0}  \\

\multicolumn{1}{|c|}{DANN~\cite{ganin2016domain}} &58.2&59.8  &61.4&62.8   &56.3&59.6  &52.8&55.4  &57.4&59.9  &52.2&54.9 &70.3&72.2 &58.4&\multicolumn{1}{c|}{60.7}  \\

\multicolumn{1}{|c|}{ADR~\cite{saito2017adversarial}} &57.1&60.7  &61.3&61.9   &57.0&60.7  &51.0&54.4  &56.0&59.9  &49.0&51.1 &72.0&74.2 &57.6&\multicolumn{1}{c|}{60.4} \\

 \multicolumn{1}{|c|}{CDAN~\cite{long2018conditional}} &65.0&69.0  &64.9&67.3   &63.7&68.4  &53.1&57.8  &63.4&65.3  &54.5&59.0 &73.2&78.5 &62.5&\multicolumn{1}{c|}{66.5} \\
 
  \multicolumn{1}{|c|}{ENT~\cite{grandvalet2005semi}} &65.2&71.0  &65.9&69.2   &65.4&71.1  &54.6&60.0  &59.7&62.1  &52.1&61.1 &75.0&78.6 &62.6&\multicolumn{1}{c|}{67.6} \\
 
\multicolumn{1}{|c|}{MME~\cite{saito2019semi}} &70.0&72.2  &67.7&69.7   &69.0&71.7  &56.3&61.8  &64.8&66.8  &61.0&61.9 &76.1&78.5 &66.4&\multicolumn{1}{c|}{68.9} \\

\multicolumn{1}{|c|}{BNM~\cite{cui2020towards}} &66.8&68.7  &67.3&68.6   &66.7&69.3  &58.2&58.3  &63.9&65.6  &59.1&60.5 &76.4&78.1 &65.5&\multicolumn{1}{c|}{67.0} \\

\multicolumn{1}{|c|}{UODA~\cite{qin2021contradictory}} &{72.7}&{75.4}  &{70.3}&{71.5} &{69.8}&{73.2}  &{60.5}&{64.1} &{66.4}&{69.4}  &{62.7}&{64.2} &{77.3}&{80.8}  &{68.5}&\multicolumn{1}{c|}{{71.2}} \\

\multicolumn{1}{|c|}{CDAC~\cite{li2021cross}} &\textbf{77.4}&\textbf{79.6}  &74.2&{75.1}  &\textbf{75.5}&\textbf{79.3}&\textbf{67.9}&{69.9}  &{71.0}&{73.4} &{69.2}&\textbf{72.5}  &\textbf{80.4}&{81.9}  &{73.6} &\multicolumn{1}{c|}{{76.0}} \\

\multicolumn{1}{|c|}{Ours} &{77.0}&{79.4}  &\textbf{75.4}&\textbf{76.7} &\textbf{75.5}&{78.3}  &{66.5}&\textbf{70.2} &\textbf{72.1}&\textbf{74.2}  &\textbf{70.9}&{72.1} &{79.7}&\textbf{82.3}  &\textbf{73.9}&\multicolumn{1}{c|}{\textbf{76.2}} \\

\hline 
\multicolumn{17}{|c|}{VGG-16}  \\\hline

   
   
\multicolumn{1}{|c|}{S+T} &49.0&52.3  &55.4&56.7   &47.7&51.0  &43.9&48.5  &50.8&55.1  &37.9&45.0 &69.0&71.7 &50.5&\multicolumn{1}{c|}{54.3}\\

\multicolumn{1}{|c|}{DANN~\cite{ganin2016domain}} &43.9&56.8  &42.0&57.5   &37.3&49.2  &46.7&48.2  &51.9&55.6  &30.2&45.6 &65.8&70.1 &45.4&\multicolumn{1}{c|}{54.7}\\

\multicolumn{1}{|c|}{ADR~\cite{saito2017adversarial}} &48.3&50.2  &54.6&56.1   &47.3&51.5  &44.0&49.0  &50.7&53.5  &38.6&44.7 &67.6&70.9 &50.2&\multicolumn{1}{c|}{53.7} \\

 \multicolumn{1}{|c|}{CDAN~\cite{long2018conditional}} &57.8&58.1  &57.8&59.1   &51.0&57.4  &42.5&47.2  &51.2&54.5  &42.6&49.3 &71.7&74.6 &53.5&\multicolumn{1}{c|}{57.2} \\
 
 \multicolumn{1}{|c|}{ENT~\cite{grandvalet2005semi}} &39.6&50.3  &43.9&54.6   &26.4&47.4  &27.0&41.9  &29.1&51.0  &19.3&39.7 &68.2&72.5 &36.2&\multicolumn{1}{c|}{51.1} \\

\multicolumn{1}{|c|}{MME~\cite{saito2019semi}} &60.6&64.1  &63.3&63.5   &57.0&60.7  &50.9&55.4  &{60.5}&60.9  &{50.2}&54.8 &72.2&75.3 &59.2&\multicolumn{1}{c|}{62.1} \\

\multicolumn{1}{|c|}{BNM~\cite{cui2020towards}} &57.3&60.3  &59.1&60.3   &54.6&59.1  &48.6&53.3  &56.1&58.4  &44.1&50.2 &71.1&73.8 &55.9&\multicolumn{1}{c|}{59.3} \\

\multicolumn{1}{|c|}{UODA~\cite{qin2021contradictory}} &{62.2}&{66.2}  &{63.6}&{65.7} &{59.4}&{65.1}  &{52.3}&{57.6} &59.2&{63.2}  &49.6&{55.9} &\textbf{74.1}&{76.3}  &{60.1}&\multicolumn{1}{c|}{{64.3}} \\

\multicolumn{1}{|c|}{Ours} &\textbf{64.9}&\textbf{67.3}  &\textbf{66.8}&\textbf{68.2} &\textbf{61.6}&\textbf{67.2}  &\textbf{55.3}&\textbf{61.1} &\textbf{63.4}&\textbf{65.5}  &\textbf{52.8}&\textbf{59.4} &{73.6}&\textbf{76.6}  &\textbf{62.6}&\multicolumn{1}{c|}{\textbf{66.5}} \\


 \hline
\end{tabular}
\renewcommand{\labelitemi}{}
\end{threeparttable}
}
\end{center}
\end{table*}

\textbf{Datasets.}  We select the latest DA benchmarks including DomainNet~\cite{peng2019moment} and Office-home~\cite{venkateswara2017deep}, for a thorough evaluation of our proposed approach. To fairly compare with the baseline methods, we use the same protocol as~\cite{saito2019semi,qin2021contradictory}. The Office-home benchmark is slightly smaller than DomainNet, which consists of 4 domains including \textit{Real} (\textbf{R}), \textit{Clipart} (\textbf{C}), \textit{Art} (\textbf{A}) and \textit{Product} (\textbf{P}) with $65$ classes shared over all the domains. We have applied all the 12 adaptation scenarios to fairly compare the proposed method with previous approaches. On the DomainNet benchmark, there are $4$ domains including \textit{Real} (\textbf{R}), \textit{Painting} (\textbf{P}), \textit{Clipart} (\textbf{C})  and \textit{Sketch} (\textbf{S}) with $126$ classes for evaluation. And there are 7 adaptation scenarios with different scales/types of domain gap to overcome. 

%
\textbf{Baselines.} \textbf{S+T} denotes the model trained by labeled source and target data only. \textbf{ENT}~\cite{grandvalet2005semi} is an SSL method that minimizes the conditional entropy on unlabeled target samples without explicit alignment. \textbf{DANN}~\cite{ganin2016domain} adversarially applies a discriminator to confuse the cross-domain features. \textbf{CDAN}~\cite{long2018conditional} employs entropy minimization to control the uncertainty of predictions for transferability. \textbf{ADR}~\cite{saito2017adversarial} is a GAN-based method designed to learn domain-invariant and discriminative features. \textbf{MME}~\cite{saito2019semi} is an SSDA method where the conditional entropy of the unlabeled target samples is adversarially minimized. \textbf{BNM}~\cite{cui2020towards} is a UDA/SSDA method attempting to maximize nuclear-norm to learn transferable features. \textbf{UODA}~\cite{qin2021contradictory} considered the source domain regularization and extended the MME based on the opposite structure learning. \textbf{CDAC}~\cite{li2021cross} is a recent paper that utilizes both adaptive adversarial clustering and unsupervised data augmentation for SSDA.

\textbf{Evaluation.} Following the protocol of ~\cite{saito2019semi}, all the baseline and proposed models will use the transductive setting where the unlabeled target samples are seen as the final target for evaluation, given the source and target labeled data and unlabeled target samples for training. All the methods will be evaluated under the one-shot and three-shot settings where there are one or three labeled samples per class in the target domain. The top-1 accuracy over all the samples has been reported as the evaluation matrix.

\textbf{Data Augmentation.} We follow the data augmentation strategies of RandAugment~\cite{cubuk2020randaugment} where two of ten augments are randomly picked for strong data transformation. The total set of the data augmentation strategies consist of $AutoContrast$, $Brightness$, $Color$, $Contrast$, $Equalize$, $Identity$, $Posterize$, $Rotate$, $Sharpness$, $ShearX$, $ShearY$, $Solarize$, $TranslateX$ and $TranslateY$. The weak augmentation is a common practice which involves $RandomHorizontalFlip$ and $RandomCrop$ from 256$\times$256 to 224$\times$224 with paddings as 28.

\subsection{Results on DomainNet.}
Table~\ref{tab:quant_domainnet} summarizes the quantitative comparison of our proposed approach with baseline methods on the DomainNet dataset. We can see that ours have outperformed all the previous methods in the average on both 1-shot and 3-shot settings. Compared with the direct baseline UODA, the current method shows a huge superiority in most of the adaptation scenarios. Especially on the ResNet-34, there is a 5\% improvement for both 1-shot and 3-shot settings. According to our survey, the most challenging domain gap to overcome is Real to Sketch, where ours have beaten the latest paper on the 1-shot setting. Since CDAC has not tested its method on the VGG-16, the strongest baseline for comparison is UODA in this case. The improvements on 1-shot SSDA are slightly inferior to those of the 3-shot on both backbones in the comparison of the performance of both tasks. This means that our methods need more supervision to better exploit its potential since more labeled target examples are helpful to indicate the conditional distributions of the target features. Our proposed approach is stably higher than UODA with the backbone of VGG-16 except for the Painting to Real scenario, where the proposed self-training and explicit brings the negative transfer. Such instability can be seen as a weakness of our method, which, however, is robust in most of the cases.

\begin{table*}[t]
\begin{center}
 \caption{{Quantitative results (\%) on Office-home~\cite{venkateswara2017deep} under the backbone of VGG-16~\cite{simonyan2014very}.} }
\label{tab:quat_office}
\scalebox{1}{
\begin{threeparttable}
 \centering
  \begin{tabular}{|cccccccccccccc|}
   \hline
\multicolumn{14}{|c|}{ONE-SHOT}  \\\hline
   \multicolumn{1}{|c|}{Methods} & {R$\rightarrow$C} & {R$\rightarrow$P} & {R$\rightarrow$A} & {P$\rightarrow$R} & {P$\rightarrow$C} & {P$\rightarrow$A} & {A$\rightarrow$P} & {A$\rightarrow$C} & {A$\rightarrow$R} & {C$\rightarrow$R} & {C$\rightarrow$A} & {C$\rightarrow$P}  & \multicolumn{1}{|c|}{{Avg}}\\\hline
\multicolumn{1}{|c|}{S+T} &39.5 &75.3 &61.2 &71.6 &37.0 &52.0 &63.6 &37.5 &69.5 &64.5 &51.4  &65.9 &\multicolumn{1}{|c|}{57.4}  \\
\multicolumn{1}{|c|}{DANN~\cite{ganin2016domain}} &\textbf{52.0} &75.7 &62.7 &72.7 &45.9 &51.3 &64.3 &44.4 &68.9 &64.2 &52.3  &65.3 &\multicolumn{1}{|c|}{60.0}  \\
\multicolumn{1}{|c|}{ADR~\cite{saito2017adversarial}} &39.7 &76.2 &60.2 &71.8 &37.2 &51.4 &63.9 &39.0 &68.7 &64.8 &50.0  &65.2 &\multicolumn{1}{|c|}{57.4} \\
 \multicolumn{1}{|c|}{CDAN~\cite{long2018conditional}} &43.3 &75.7 &60.9 &69.6 &37.4 &44.5 &67.7 &39.8 &64.8 &58.7  &41.6 &66.2 &\multicolumn{1}{|c|}{55.8} \\
  \multicolumn{1}{|c|}{ENT~\cite{grandvalet2005semi}} &23.7 &77.5 &64.0 &74.6 &21.3 &44.6 &66.0 &22.4 &70.6 &62.1  &25.1 &67.7 &\multicolumn{1}{|c|}{51.6} \\
 
\multicolumn{1}{|c|}{MME~\cite{saito2019semi}} &49.1 &78.7 &65.1 &74.4 &{46.2} &56.0 &68.6 &\textbf{45.8} &72.2 &68.0 &57.5  &71.3 &\multicolumn{1}{|c|}{62.7} \\

\multicolumn{1}{|c|}{BNM~\cite{cui2020towards}} &51.0 &79.5 &62.8 &72.3 &44.0 &51.8 &67.1 &45.7 &68.4 &65.3 &52.7  &69.1 &\multicolumn{1}{|c|}{60.8} \\

\multicolumn{1}{|c|}{UODA~\cite{qin2021contradictory}} &49.6 &{79.8}  &{66.1} &{75.4} &{45.5} &{58.8} &\textbf{72.5} &{43.3} &\textbf{73.3} &{70.5} &\textbf{59.3} &{72.1}  &\multicolumn{1}{|c|}{{63.9}} \\

\multicolumn{1}{|c|}{Ours} &{51.6} &\textbf{80.9}  &\textbf{66.9} &\textbf{75.9} &\textbf{49.7} &\textbf{60.5} &{71.0} &{44.9} &{73.2} &\textbf{70.6} &{58.7} &\textbf{72.8}  &\multicolumn{1}{|c|}{\textbf{64.7}} \\

     \hline 
     
 \multicolumn{14}{|c|}{THREE-SHOT}  \\\hline
\multicolumn{1}{|c|}{S+T} &49.6 &78.6 &63.6 &72.7 &47.2 &55.9 &69.4 &47.5 &73.4 &69.7 &56.2  &70.4 &\multicolumn{1}{|c|}{62.9}  \\
\multicolumn{1}{|c|}{DANN~\cite{ganin2016domain}} &56.1 &77.9 &63.7 &73.6 &52.4 &56.3 &69.5 &50.0 &72.3 &68.7 &56.4  &69.8 &\multicolumn{1}{|c|}{63.9}  \\
\multicolumn{1}{|c|}{ADR~\cite{saito2017adversarial}} &49.0 &78.1 &62.8 &73.6 &47.8 &55.8 &69.9 &49.3 &73.3 &69.3 &56.3  &71.4 &\multicolumn{1}{|c|}{63.0} \\
 \multicolumn{1}{|c|}{CDAN~\cite{long2018conditional}} &50.2 &80.9 &62.1 &70.8 &45.1 &50.3 &74.7 &46.0 &71.4 &65.9  &52.9 &71.2 &\multicolumn{1}{|c|}{61.8} \\
  \multicolumn{1}{|c|}{ENT~\cite{grandvalet2005semi}} &48.3 &81.6 &65.5 &76.6 &46.8 &56.9 &73.0 &44.8 &75.3 &72.9  &59.1 &77.0 &\multicolumn{1}{|c|}{64.8} \\
 
\multicolumn{1}{|c|}{MME~\cite{saito2019semi}} &56.9 &82.9 &65.7 &76.7 &53.6 &59.2 &75.7 &54.9 &75.3 &72.9 &61.1  &76.3 &\multicolumn{1}{|c|}{67.6} \\

\multicolumn{1}{|c|}{BNM~\cite{cui2020towards}} &56.0 &81.2 &64.7 &73.8 &52.7 &55.8 &72.1 &51.7 &73.2 &70.3 &57.0  &73.3 &\multicolumn{1}{|c|}{65.2} \\

\multicolumn{1}{|c|}{UODA~\cite{qin2021contradictory}} &{57.6} &{83.6}  &{67.5} &{77.7} &{54.9} &{61.0} &{77.7} &{55.4} &\textbf{76.7} &{73.8} &{61.9} &{78.4}  &\multicolumn{1}{|c|}{{68.9}} \\

\multicolumn{1}{|c|}{Ours} &\textbf{59.3} &\textbf{83.6}  &\textbf{68.0} &\textbf{78.3} &\textbf{56.8} &\textbf{61.8} &\textbf{78.6} &\textbf{55.7} &{75.3} &\textbf{74.0} &\textbf{63.3} &\textbf{78.9}  &\multicolumn{1}{|c|}{\textbf{69.5}} \\

\hline
\end{tabular}
\end{threeparttable}
}
\end{center}
\end{table*}

\begin{figure*}[t]
\centering
\scalebox{1}{\includegraphics[width=1.0\textwidth]{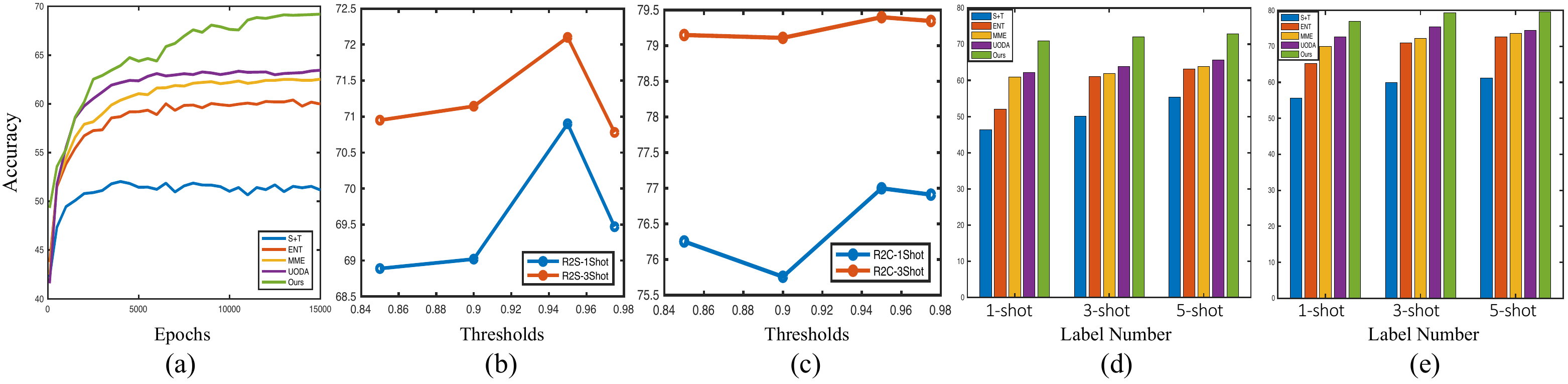}}
\caption{(a): Convergence analysis of four baselines and ours on R$\rightarrow$S. (b) - (c): The performance of our method in different values of threshold $\tau$ (with $\tau$=0.85, $\tau$=0.9, $\tau$=0.95, $\tau$=0.975) on {R$\rightarrow$S}, i.e., sub-figure (b) and {R$\rightarrow$C}, i.e., sub-figure (c). (d) - (e): Histogram of  quantitative comparisons under 1-shot, 3-shot, and 5-shot settings on R$\rightarrow$S, i.e., sub-figure (d) and R$\rightarrow$C, i.e., sub-figure (e). 
} \label{fig:joint_sub_figs}
\end{figure*}

\begin{figure*}[t]
\centering
\scalebox{1}{\includegraphics[width=0.9\textwidth]{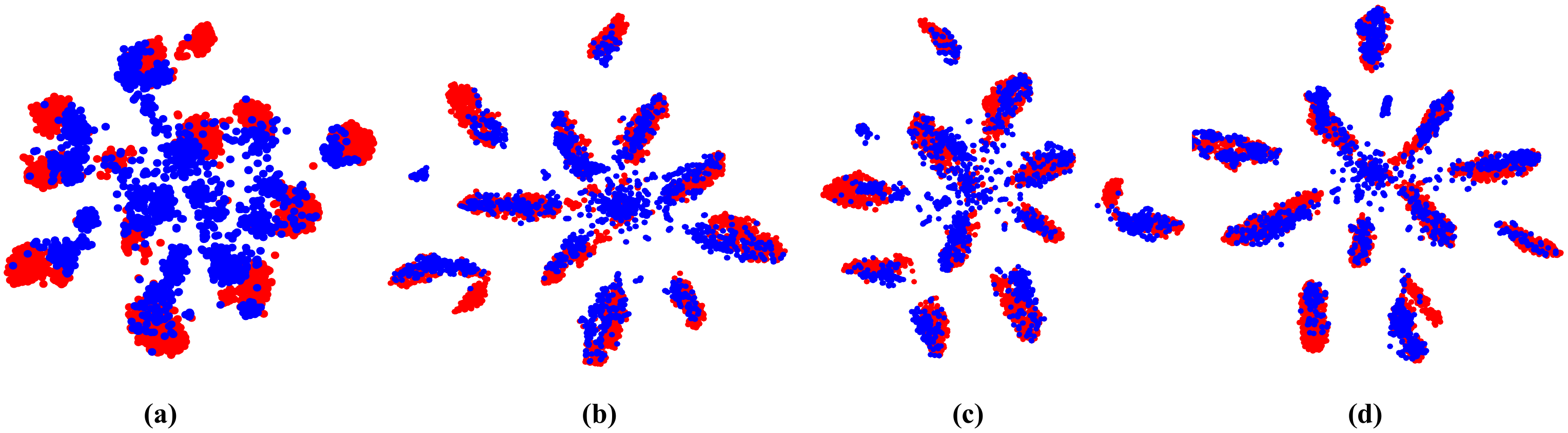}}
\caption{ The t-SNE~\cite{maaten2008visualizing} visualization results of shared top ten-class features in the 3-shot R$\rightarrow$S problem  obtained by: (a) S+T, (b) MME, (c) UODA, (d) Ours. The red and blue spots indicate the source and target feature points, respectively.
} \label{fig:tsne}
\end{figure*}

\subsection{Results on Office-Home.}
Table~\ref{tab:quat_office} shows the quantitative results and comparison on the benchmark Office-home. We can notice that the proposed approach has outperformed the direct baseline, i.e., UODA, on most adaptation scenarios. However, on the 1-shot setting, ours is largely beaten by the UODA on the case of adapting from Clipart to Art. This likely happens when the domain gap is large with the diversified data that self-training will cause more substantial confirmation bias towards the labeled target set. Moreover, on the scenarios as Art to Real and Clipart to Real, the UODA and ours have similar results indicating that the proposed self-training and explicit alignment cause no extra boost or damage for the transfer. On the 3-shot setting, ours is also inferior to UODA on the Art to Real scenario, which shows the similar phenomenon as the 1-shot setting. Compared with the results on DomainNet, the improvements on Office-home are significantly weaker in both 1-shot and 3-shot settings despite the fact that the Office-home dataset seems more straightforward to solve due to fewer classes. This may be explained as the more effective self-training in DomainNet due to the larger quantity of unlabeled data for the regularizing more smoothing decision boundaries.

\begin{table*}[t]
\begin{center}
\caption{ Quantitative results (\%) of ablation study by the backbone ResNet34~\cite{he2016deep}.}\label{tab:ablation}
\scalebox{0.99}{
\begin{threeparttable}
 \centering
  \begin{tabular}{|cccccc|cc|cc|}
\hline 
\multicolumn{6}{|c}{{COMPONENTS}}&\multicolumn{2}{c}{{REAL $\longrightarrow$ SKETCH}}&\multicolumn{2}{c|}{{REAL  $\longrightarrow$  CLIPART}}\\
\multicolumn{1}{|c}{~1-$C$~} &\multicolumn{1}{c}{~2-$C$~}  &\multicolumn{1}{c}{~${H}_{tar}$~} &\multicolumn{1}{c}{${H}_{tar}$+${H}_{src}$}  &\multicolumn{1}{c}{MMD} &\multicolumn{1}{c}{~~ST~~}&ONE-{SHOT} & \multicolumn{1}{c}{THREE-{SHOT}}
&\multicolumn{1}{c}{ONE-{SHOT}} & THREE-{SHOT} \\
\hline
{{\Checkmark}} &{ } &{ \Checkmark} &{ } &{ } &{ }  &{61.03}&61.93  &70.04&72.19\\

{\Checkmark} &{ } &{ }  &{\Checkmark } &{ } &{ }  &60.40&61.16  &69.24&71.47\\


{\Checkmark } &{} &{\Checkmark}  &{ } &{\Checkmark } &{\Checkmark}  &\textbf{69.39}&\textbf{71.03} &\textbf{73.94}&\textbf{77.88}\\

\hline

{ } &{\Checkmark} &{\Checkmark } &{ } &{ } &{ }  &61.11&62.81  &70.53&72.42\\

{ } &{\Checkmark} &{ }  &{ \Checkmark} &{ } &{ }  &62.17&63.90  &71.57&74.02\\

{ } &{\Checkmark} &{ }  &{ \Checkmark} &{ } &{\Checkmark}  &{69.16}&{71.46}  &{75.68}&{78.57}\\

{ } &{\Checkmark} &{ }  &{ \Checkmark} &{ \Checkmark} &{}  &65.58&67.18  &74.11&75.34\\

{} &{\Checkmark } &{ }  &{ \Checkmark} &{\Checkmark } &{\Checkmark}  &\textbf{70.9}1&\textbf{72.12}  &\textbf{76.98}&\textbf{79.40}\\


\hline 
\end{tabular}
\renewcommand{\labelitemi}{}
\end{threeparttable}
}
\end{center}
\end{table*}

\subsection{Ablation Study.}
\label{sec:analysis}
Since the proposed system relies on multiple components for fulfilling the final objective, it is necessary to fully investigate their effects. As shown in the Table~\ref{tab:ablation}, there is a detailed ablation study for the analysis of each component in the four transfer scenarios, including Real to Sketch and Real to Clipart on both 1 or 3-shot settings.  Six modules are chosen for analysis including the model with one classifier ({1-{$C$}}) or two classifiers ({2-{$C$}}) composed of target-clustering classifier and source-scattering classifier,  the target entropy ({${H}_{tar}$}), and the combination of the source and target entropy ({${H}_{tar}$}+{${H}_{src}$}), and the explicit alignment loss ({MMD}) and  
 self-training ({ST}). 
 
In the top three rows, which compare the results of one classifier structure, the explicit alignment and self-training bring significant improvements on all the adaptation cases. Furthermore, the contradictory structure learning on one classifier will, however, cause a performance drop, indicating the necessity of two classifiers. The bottom five rows summarize the results with two classifiers as utilized in our method. The effectiveness of contradictory structure learning can be verified by comparing row 4 and row 5, with or without the source entropy loss. Over all the scenarios, the latter is obviously inferior to the former with the contradictory structure learning. Finally, compared with the MMD only (row 6) and ST (row7) only cases, it is easy to notice that the self-training contributes more to the performance boost. And the joint applying of MMD and ST will achieve the best as expected.

    

\begin{table}[t]
\begin{center}
 \caption{Split Analysis of Real to Sketch on Three Shot Setting }
\label{tab:split}
\scalebox{1.1}{
\begin{threeparttable}
 \centering
  \begin{tabular}{|c|ccc|c|}
\hline 
{Methods} &{Split-1} &{Split-2} &{Split-3} &{Avg $\pm$ Var}  \\\hline

{S+T} &50.1  &52.7 &51.5  &51.4 $\pm$ 1.7\\
{MME~\cite{saito2019semi}} &61.9  &61.2 &63.8 &62.3 $\pm$ 1.8\\
{UODA~\cite{qin2021contradictory}} &64.2  &64.1 &63.0 &63.8 $\pm$ 0.4 \\
{Ours} &\textbf{72.1}  &\textbf{72.1} &\textbf{70.9} &\textbf{71.7 $\pm$ 0.5} \\
\hline 
\end{tabular}
\end{threeparttable}
}
\end{center}
\end{table}

\subsection{Analyses.}
\textbf{Convergence Analysis. } The Fig.~\ref{fig:joint_sub_figs} (a) shows the accuracy along with the training of different methods, including both baselines and ours for the 1-shot R$\rightarrow$S problem. It is clear to notice that the accuracy of the proposed method has continuously improving whereas the others are converged in the early stages. The performance gap keeps growing through the iterations, which is mostly contributed from the self-training and explicit alignment. More high-confident pseudo labels are generated due to the increase of softmax scores during training.

\textbf{Sensitivity of Threshold $\tau$.} 
As shown in Fig.~\ref{fig:joint_sub_figs} (b) - (c), we have analyzed the sensitivity of threshold $\tau$ on both Real to Sketch and Real to Clipart settings with the options as \{$\tau$=0.85, $\tau$=0.9, $\tau$=0.95, $\tau$=0.975\}. We can clearly observe that the proposed approach will achieve the best when setting the threshold as 0.95, which is consistent to the SSL methods~\cite{sohn2020fixmatch}.

\textbf{Sensitivity of labeled samples. } In the sub-figure (d) - (e) of  Fig.~\ref{fig:joint_sub_figs}, we have shown the histogram for comparisons with different labeled samples. We can draw similar conclusions from both sub-figures that our proposed method has achieved the top results despite the number of labeled data. However, the improvements from 3-shot to 5-shot are slightly narrow than those of 1-shot to 3-shot. Such a phenomenon indicates the diminishing gains from more labeling, which will eventually converge to the fully supervised one.

\subsection{Feature Visualization.}
As shown in Fig.~\ref{fig:tsne}, we take the t-SNE~\cite{maaten2008visualizing} to reduce the dimensions of raw deep features into two for visualization. Compared with the feature map of S+T, the other three adaptive methods have more dense and separate features, which obviously are better aligned. Although the difference between the MME, UODA, and ours is not significant, we can still notice that the clusters are more uniformly distributed and more separable, leading to more accurate transfer.

\subsection{Split Analysis.} 
To verify the robustness of the proposed method over the baselines against the data selections/bias, Table~\ref{tab:split} shows the multiple results on three different splits of labeled/unlabeled target domain data. Ours has achieved the best on all the splits with a relatively low variance to show good stability against the bias of selected labeled data.

\section{Conclusion}
Semi-supervised domain adaptation (SSDA) is crucial in transfer learning which has a significant performance boost compared to the unsupervised domain adaptation with only marginal annotations. This paper attempts to address the SSDA by learning the adaptive structures for feature transfer. The previous implicit feature alignment for learning well-clustered target features and scattered source features may result in the categorical mismatch across domains. To solve this, we have applied explicit alignment by minimizing the distance (i.e., MMD loss) between pairs of cross-domain features in the reproducing kernel Hilbert space. It helped to project the contradictory structures into a shared view for the robust final decision. Moreover, pseudo-labeling is employed to regularize the decision boundary towards smoothness in a self-training manner. Extensive experiments on the multiple benchmarks, including Office-home~\cite{venkateswara2017deep} and DomainNet~\cite{peng2019moment}, have shown the advantages of our proposed approach over our direct baseline~\cite{qin2021contradictory} and other latest methods. Our work may further inspire the community to investigate the better adaptive structures for transfer learning.


\ifCLASSOPTIONcaptionsoff
  \newpage
\fi


%


\bibliographystyle{IEEEtran}
\bibliography{reff, ref_new}

\end{document}